\documentclass[10pt]{article}
\usepackage[preprint]{tmlr}

\usepackage{amsmath,amssymb,amsthm}
\usepackage{booktabs}
\usepackage{float}
\usepackage{graphicx}
\usepackage{tabularx}
\usepackage{hyperref}
\usepackage{url}
\hypersetup{hidelinks}

\newtheorem{proposition}{Proposition}
\newtheorem{lemma}{Lemma}
\newcommand{\E}{\mathbb{E}}
\newcommand{\R}{\mathbb{R}}
\newcommand{\KL}{\operatorname{KL}}
\newcommand{\tr}{\operatorname{tr}}
\newcommand{\diag}{\operatorname{diag}}

\title{When Compression Scores Cannot Decide:\\
Information Boundaries for Group-Robust LLM Pruning}

\author{\name Andrew Zhang \email{yibingz@kth.se} \\
\addr KTH Royal Institute of Technology}

\begin{document}

\maketitle

\begin{abstract}
A stable compression score can still select the worse model. In our dense
study, a split-half reliable path-quadratic score predicted a 16.1\% gain,
while the selected endpoints were 6.0--7.7\% worse than two controls. We ask
what a compression statistic can justify when deployment cares about the worst
supplied group. We treat each statistic as an information interface. Its
observation leaves a fiber of compatible endpoint-risk tables, and only orders
fixed across that fiber are identified. Cone and fiber identities quantify the
remaining uncertainty, while matched observations reverse endpoint order for
pooled moments, group-local moments, and reference-path curvature. Sequential
composition adds one state variable: the slack from each group risk to the
current maximum. This vector determines every unrestricted one-step response,
and a margin condition keeps the active group fixed along paths with bounded
relative drift. The experiments follow the same ladder. Across three dense
LLMs, an early-preserving allocation reduces worst-group perplexity inflation
by 12.6--20.9\%; target-matched complete-menu selection improves over its
references by 2.7--8.0\%. Across all 16 routed layers of OLMoE, pooled endpoint
refresh lowers held-out worst-group teacher KL by 15.8\% over the best static
score. A compute-matched hard-max trajectory ends 32.7\% worse than pooled,
and neither adaptive trajectory improves excess NLL. Local evidence can narrow
a menu. Complete endpoints rank that menu, while multistep claims also require
control of the evolving active face and future candidates.
\end{abstract}

\section{Introduction}

One-shot methods such as Wanda and SparseGPT remove much of a language model
after a short calibration pass \citep{sun2024wanda,frantar2023sparsegpt}.
Their scores average local reconstruction evidence over a calibration
distribution, whereas deployment may ask whether every supplied group remains
covered. A direction concentrated in one group can look harmless in the
average and dominate that group's endpoint damage.
Related evaluations report subgroup loss, input-level quantization failures,
and broader trustworthiness regressions under compression
\citep{gee2023compressed,chang2025qerror,hong2024compressedtrust}.

Group resolution alone does not settle the choice. In our dense study, a
highly reproducible path-quadratic score predicts gains near 16\% and selects
endpoints that lose to both controls. The score measures local geometry along a
reference path. The decision compares complete sparse models away from that
path. Therefore, this is an information mismatch before it is an optimization problem.

\begin{figure}[H]
\centering
\includegraphics[width=\linewidth]{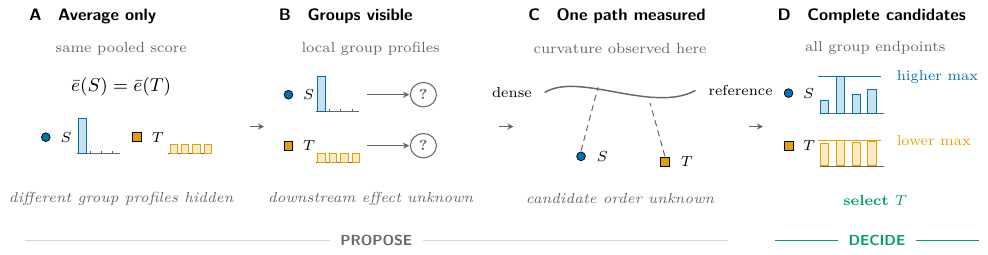}
\caption{Interfaces authorize different decisions. An average can hide group
structure (A); group-resolved local evidence omits downstream continuation
(B); and reference-path curvature does not uniformly determine off-path order (C).
Measuring the paired group-endpoint contrast reveals the displayed
order (D). Panels A through C propose; panel D decides this comparison.}
\label{fig:theory-witnesses}
\end{figure}

We treat a compression statistic as an \emph{information interface}. The
\emph{boundary between proposal and decision} marks the distinctions that remain identifiable
from that interface. Pooled moments see an average. Group-resolved moments see
the rows of that average, while downstream continuation remains hidden. A path
score sees local geometry along the path it measures. Sparse MoE routers expose
a richer interface because their traces attach group information directly to
removable experts. Figure~\ref{fig:theory-witnesses} gives the intuition before
the formal development.

Two admissible worlds may expose the same observation and reverse the endpoint
order. The compatible endpoint tables form an \emph{observation fiber}, whose
radius measures the information left unresolved. We now formalize this object
and examine three interfaces through it.

The theory and experiments follow the same ladder. Group-resolved information
recovers broad severity, while local evidence loses fine endpoint order.
Complete-candidate measurements recover gains on finite dense and MoE menus.
Multi-step composition adds the slack from current group risks to the
active-group chamber walls. Cone, minimax, and max-plus arguments quantify
these static and sequential boundaries.

\section{Compression as an information-interface problem}

Figure~\ref{fig:theory-witnesses} ends with the decision of interest, the
endpoint order of two complete candidates. That endpoint is the natural
starting point. The cheaper quantities below are partial observations of it.

Let $z$ collect latent quantities that can affect endpoint risk, and let
$\mathsf I(z)=o$ denote the observation exposed by an interface. If
$\mathsf Y(z)$ is the candidate-by-group endpoint field, the framework follows
the chain
\begin{equation}
o\ \longmapsto\
\mathcal C_o=\{z:\mathsf I(z)=o\}\ \longmapsto\
\mathcal F_o=\{\mathsf Y(z):z\in\mathcal C_o\}\ \longmapsto\
d(\mathcal F_o).
\label{eq:authorization-chain}
\end{equation}
Here $d$ is the scalar contrast used by the decision, and
$d(\mathcal F_o)$ is its set of compatible values. The interface identifies
the order exactly when this set lies strictly on one side of zero. The compared
intervention family and evaluation horizon specify the decision grain.

Let the dense teacher have parameters $W$ and let $W_M=W\odot(1-M)$ be a
frozen-weight masked student. Two group endpoints recur in the paper. For a
sequence $x$ from supplied group $g$, the label-free function-coverage endpoint
is
\begin{equation}
J_g(M)=\E_{x\sim P_g}\left[
\frac{1}{L_x}\sum_{t=1}^{L_x}
\KL\bigl(p_{W,t}(\cdot\mid x_{<t})\,\|\,
p_{W_M,t}(\cdot\mid x_{<t})\bigr)\right].
\label{eq:endpoint-kl}
\end{equation}
Write $\mathcal J(M)=\max_gJ_g(M)$. Given a declared token loss $\ell$, the
compression-performance endpoint is
\begin{equation}
E_g(M)=\E_{x\sim P_g}[\ell(W_M;x)-\ell(W;x)],
\qquad \mathcal E(M)=\max_gE_g(M).
\label{eq:endpoint-excess}
\end{equation}
Forward KL measures teacher-to-student coverage on the supplied support; $E_g$
measures the loss caused by compression. A guarantee for one endpoint does not
in general transfer to the other. We use $F_g$ and $\mathcal T_F(M)=\max_gF_g(M)$ when a
statement holds for either declared endpoint.

Both endpoints are measurable for a fixed candidate. The difficulty is the
cost of candidate-specific queries. Dense pruning reduces that cost by
generating a finite menu of complete masks; MoE exposes a smaller menu of
expert interventions.

To see what the cheaper interfaces lose, Panels A and B use a simpler local
account. Removing unit $u$ contributes a nonnegative amount to each supplied
group $g$. For removable units
$u\in\mathcal U$ and groups $g\in[G]$, collect these contributions in
$\mathbf e=(e_{ug})_{u,g}$ and write
\begin{equation}
e_{ug}\geq0,
\qquad
D_g(S)=\sum_{u\in S}e_{ug},
\label{eq:energy}
\end{equation}
for a removed set $S$. In a dense linear layer,
$e_{ug}=w_u^2n_g(u)^2$, where $n_g(u)^2$ is the group-conditional input second
moment seen by the weight. This is the diagonal reconstruction surrogate. It is
exact when the corresponding cross terms vanish and otherwise remains the
declared local approximation.
The conic pooling identity in Equation~\ref{eq:conic-pooling-identity} extends
the same factors to full PSD input moments for a fixed candidate, although
off-diagonal terms prevent reusable per-weight scores. Both accounts remain
layer local. Section~\ref{sec:moe-interface}
uses the same nonnegative interface to measure routed-expert exposure.

For finitely many nonempty groups with equal composition, write
$\bar e_u=G^{-1}\sum_g e_{ug}$ and
$D_{\mathrm{mix}}(S)=\sum_{u\in S}\bar e_u$. A pooled interface retains only
$(\bar e_u)_u$, a group-resolved interface retains $(e_{ug})_{u,g}$, and a
capped interface also declares $e_{ug}\leq c\bar e_u$. The next section
formalizes these cases once for a general positive cone.

Table~\ref{tab:notation} collects the recurring notation. Symbols introduced
for a single proof remain local to that proof.

\begin{table}[H]
\centering
\small
\caption{Recurring notation, grouped by the interface that supplies it.}
\label{tab:notation}
\begin{tabularx}{\linewidth}{@{}l l X@{}}
\toprule
Interface & Symbols & Meaning \\
\midrule
Framework & $\mathsf I,\mathcal C_o,\mathcal F_o,d$ & observation map, compatible worlds, endpoint fiber, and decision contrast \\
Endpoint & $J_g,E_g,F_g,\mathcal T_F$ & coverage, performance, a generic group endpoint, and its worst-group aggregate \\
Local damage & $D_g,D_{\mathrm{mix}}$ & group damage and its pooled counterpart \\
Relaxation & $\Phi_D,\Psi_D$ & exact mask maximum and per-unit switching relaxation \\
Exposure & $\rho,\lambda$ & group contrast and restoration gain per recovered mass \\
Composition & $r,\xi,s,q_r$ & group-risk state, increment, slack, and one-step response \\
Router & $A_{\ell eg},Q_{\ell eg}$ & route mass and contribution energy \\
MoE endpoint & $\widetilde J,\widetilde I$ & normalized singleton damage and pair interaction \\
\bottomrule
\end{tabularx}
\end{table}

\section{What an information interface can authorize}
\label{sec:averaging-costs}

Panel A of Figure~\ref{fig:theory-witnesses} compares the profiles
$(4,0,0,0)$ and $(1,1,1,1)$. Both have mean one, while their largest entries are
four and one. If the mean is all that remains visible, an admissible
decomposition can concentrate the total mass in one group and produce a factor
of $G$. The same construction concentrates mass in the full PSD second-moment
case. Its common structure is a positive linear damage functional on a cone.
We now state that structure once.

Let $\mathcal K$ be a convex cone containing zero in a real vector space,
$\bar\theta\in\mathcal K$, and
write $\theta\preceq_{\mathcal K}\theta'$ when
$\theta'-\theta\in\mathcal K$. If $L_S$ is linear and nonnegative on
$\mathcal K$, then for $G\geq2$ and $1\leq c\leq G$,
\begin{align}
\sup_{\substack{\theta_g\in\mathcal K\\
G^{-1}\sum_g\theta_g=\bar\theta}}
\max_gL_S(\theta_g)
&=G L_S(\bar\theta),\nonumber\\
\sup_{\substack{\theta_g\in\mathcal K,\;
G^{-1}\sum_g\theta_g=\bar\theta\\
\theta_g\preceq_{\mathcal K}c\bar\theta\;\forall g}}
\max_gL_S(\theta_g)
&=c L_S(\bar\theta).
\label{eq:conic-pooling-identity}
\end{align}
We call this the \emph{conic pooling identity}. Positivity supplies the upper
bounds, and the concentrated profiles
$(G\bar\theta,0,\ldots,0)$ and
$(c\bar\theta,(G-c)\bar\theta/(G-1),\ldots)$ attain them. The case $G=1$ is
immediate.

For the diagonal account, take $\mathcal K=\R_+^{|\mathcal U|}$,
$\bar\theta=(\bar e_u)_u$, $\theta_g=(e_{ug})_u$, and
$L_S(\theta)=\sum_{u\in S}\theta_u$. The worst compatible values at the pooled,
group-resolved, and capped interfaces are therefore
\begin{equation}
GD_{\mathrm{mix}}(S),\qquad
\max_gD_g(S),\qquad
cD_{\mathrm{mix}}(S),
\label{eq:exact-interface-values}
\end{equation}
respectively.
Full PSD second moments take $\mathcal K$ to be the PSD cone and
$L_S(\Sigma)=\tr(\Delta W_S\Sigma\Delta W_S^\top)$, where $\Delta W_S$
is the weight perturbation induced by candidate $S$. Cone order then becomes
Loewner order and the same factors remain exact. Thus $G$ measures the
uncertainty introduced by discarding the decomposition, while $c$ measures
what remains after a declared concentration cap. Diagonalization supplies
reusable per-weight scores; the pooling factors themselves do not require it.

Per-unit group-max sorting optimizes a separable relaxation in which the
maximizing group may change from one removed unit to the next. The complete-mask
objective keeps one group fixed across the sum. Appendix~\ref{app:switching}
gives the resulting switching premium and its equality condition.

\subsection{One obstruction at three interfaces}

Panels A, B, and C hide different objects, yet their order failures share one
argument. If two admissible worlds expose the same observation and reverse two
equal-budget candidates, every deterministic rule that reads only that
observation fails in one world. Figure~\ref{fig:theory-witnesses} depicts the
three observations in the following proposition.

\begin{proposition}[Three matched-observation reversals]
\label{prop:matched-reversals}
Each interface below admits two worlds with the same observation and opposite
strict orders over two equal-budget candidates.
\begin{enumerate}
\item Two group decompositions can share every pooled row mean and reverse the
exact worst-group mask order.
\item For every $0<\mu<\Lambda$, fixed input-side quantities $(W,\Sigma)$ and
two equal-budget masks admit positive-definite downstream pullbacks with spectra
in $[\mu,\Lambda]$ that reverse the mask order.
\item For a compact line segment $\Gamma$ and distinct candidate endpoints
$a,b\notin\Gamma$ in a finite-dimensional real parameter space, two
nonnegative $C^\infty$ losses can have identical derivatives of every order on
$\Gamma$ and opposite orders at $a$ and $b$.
\end{enumerate}
\end{proposition}

\paragraph{Pooled decomposition.}
The first witness swaps the profiles $(4,0,0,0)$ and $(1,1,1,1)$ between two
units. Their pooled means agree, while their worst-group values are four and
one. For a fixed cardinality budget $|S|=k$, selecting the $k$ smallest
$\bar e_u$ exactly minimizes the pooled value in
Equation~\ref{eq:exact-interface-values}. The rule is optimal for the pooled
interface, whose observation still cannot identify group order.

\paragraph{Downstream continuation.}
Panel B reveals the local group profiles and leaves their downstream effect
unresolved. Let $\Sigma_g=\E_{x\sim P_g}[xx^\top]$ be the group input
second-moment matrix at the layer. For the weight perturbation $\Delta W_S$,
define
\begin{equation}
K_g(S)=\Delta W_S\Sigma_g\Delta W_S^\top,
\qquad
\mathcal R^C(S)=\max_g\tr(C_gK_g(S)),\quad C_g\succeq0.
\label{eq:downstream-risk}
\end{equation}
The input-side interface sees $K_g$ and implicitly prices its output directions
uniformly. It does not observe $C_g$. The witness takes
$W=\Sigma=I_2$, removes opposite output coordinates, and swaps
$\diag(\Lambda,\mu)$ with $\diag(\mu,\Lambda)$. Uniform dominance over PSD
pullbacks requires Loewner dominance of the output-error matrices.

\paragraph{Reference chord.}
Panel C measures the reference segment $\Gamma$ while both candidate endpoints
remain off it. Smooth bumps supported near those endpoints vanish to every
order on $\Gamma$ and reverse their endpoint values. A reference-path argument
therefore needs candidate coverage, a uniform remainder bound, or endpoint
measurements.

Appendix~\ref{app:interface-proofs} gives the three constructions and positive
escape conditions.

\subsection{The exact price of unresolved endpoints}

The downstream witness also gives a bounded example. For fixed $K\succeq0$
and $\mu I\preceq H\preceq\Lambda I$, the same observed $K$ permits risks from
$\mu\tr(K)$ to $\Lambda\tr(K)$. Any single prediction misses one end of this
interval by at least half its width. Across a finite candidate-by-group table,
the endpoint-risk fields compatible with observation $o$ form an
\emph{observation fiber} $\mathcal F_o$.
For a nonempty bounded fiber in $\R^{\mathcal I}$, its exact minimax prediction
radius under $\ell_\infty$ loss is
\begin{equation}
\inf_{a\in\R^{\mathcal I}}\sup_{y\in\mathcal F_o}\|y-a\|_\infty
=\frac12\operatorname{diam}_\infty(\mathcal F_o).
\label{eq:fiber-radius}
\end{equation}
The coordinatewise midrange attains the infimum. This follows by taking the
midpoint of each coordinate interval; limiting sequences give the same lower
bound when an extremum is not attained. For the spectral continuation class,
the exact price is $(\Lambda-\mu)\tr(K)/2$.

The full endpoint field can remain ambiguous even when one comparison is
settled. For a scalar decision contrast $d:\mathcal F_o\to\R$ with bounded
image, set
$d_-:=\inf_{y\in\mathcal F_o}d(y)$ and
$d_+:=\sup_{y\in\mathcal F_o}d(y)$. Its minimax prediction error is
$(d_+-d_-)/2$, and its sign is identified exactly when $[d_-,d_+]$ excludes
zero. If a common estimate $\widehat d$ has simultaneous error at most $\eta$,
the corresponding sufficient tests are
$\widehat d-\eta>0$ and $\widehat d+\eta<0$.

An unbounded fiber, including the unrestricted
off-chord construction, yields no finite uniform certificate. The next result
states what a uniform error bound buys.

\subsection{Uniform control authorizes threshold and selection decisions}

Panel D shows one comparison between complete candidates. Sweeping the
tolerance produces nested feasible sets. Let $\mathcal X$ be a finite queried
mask family and, for a scalar score $u$, define
$\mathcal X_t^u=\{M\in\mathcal X:u(M)\leq t\}$.
Write $f(M)=\psi(F_1(M),\ldots,F_G(M))$ and
$\widehat f(M)=\psi(\widehat F_1(M),\ldots,\widehat F_G(M))$ for a declared
endpoint family $F_g$.

\begin{proposition}[What uniform control authorizes]
\label{prop:uniform-control}
Suppose $\varepsilon\geq0$ and
\begin{equation}
\sup_{M\in\mathcal X}\max_g
|F_g(M)-\widehat F_g(M)|\leq\varepsilon
\label{eq:uniform-interface-error}
\end{equation}
and $\psi$ is $L_\psi$-Lipschitz from $\ell_\infty^G$ to $\R$ for some
$L_\psi\geq0$. Every threshold set moves by at most
$L_\psi\varepsilon$. If $\widehat M$ minimizes $\widehat f$ on $\mathcal X$, then
\begin{equation}
f(\widehat M)\leq\min_{M\in\mathcal X}f(M)+2L_\psi\varepsilon.
\label{eq:filtration-selection-regret}
\end{equation}
\end{proposition}

Panel D is the finite-menu, $\varepsilon=0$ case. With approximate endpoints,
uniform error moves each feasible threshold by at most
$L_\psi\varepsilon$. Both $\max_g$ and the normalized mean are 1-Lipschitz in
$\ell_\infty$, so this worst-case selection guarantee follows from simultaneous
control over the full menu. Without such uniform control, the proposition
licenses only explicitly measured finite menus.

\subsection{Sequential composition needs the active-face state}

A finite-menu guarantee ends when one candidate is accepted. The next menu is
evaluated from a new risk state, and two states with the same current maximum
can respond differently to the same increment. For a group-risk vector
$r\in\R^G$ and a possible increment $\xi\in\R^G$, define
\begin{equation}
m(r)=\max_g r_g,
\qquad s_g(r)=m(r)-r_g,
\qquad q_r(\xi)=m(r+\xi)-m(r).
\label{eq:response-state}
\end{equation}
The maximum partitions risk space into chambers indexed by the active group.
The slack vector records the distances, along group coordinates, to the walls
of those chambers. Direct substitution gives
\begin{equation}
q_r(\xi)=\max_g\{\xi_g-s_g(r)\}.
\label{eq:slack-adjusted-change}
\end{equation}

The one-step response has an exact quotient description:
\begin{equation}
q_r=q_{r'}\ \text{on }\R^G
\quad\Longleftrightarrow\quad
s(r)=s(r')
\quad\Longleftrightarrow\quad
r'=r+c\mathbf 1\ \text{for some }c\in\R.
\label{eq:response-state-characterization}
\end{equation}
Thus two profiles have the same one-step response exactly when they differ by
a common shift, the familiar max-plus projective quotient
\citep{develin2004tropical}. In compression terms, the slack vector contains
all information needed by the unrestricted one-step response map. It remains sufficient on every
restricted menu, although a particular menu may require less information.
After an increment, it updates exactly as
\begin{equation}
s_g(r+\xi)=q_r(\xi)-\{\xi_g-s_g(r)\}.
\label{eq:slack-state-update}
\end{equation}
For a candidate $M$ relative to $M_{\mathrm{ref}}$, set
$r_g=F_g(M_{\mathrm{ref}})$ and
$\xi_g=\delta_g=F_g(M)-F_g(M_{\mathrm{ref}})$. Equation
\ref{eq:slack-adjusted-change} says that $M$ strictly improves the endpoint
exactly when $\delta_g<s_g(r)$ for every group.

A simple margin condition recovers a safe horizon. Let $g^\star$ be the unique
maximizer of the initial profile. Suppose the relative increment at step $t$
satisfies $\xi_{t,g}-\xi_{t,g^\star}\leq b_{t,g}$ with $b_{t,g}\geq0$. If
\begin{equation}
\sum_{t=0}^{h-1}b_{t,g}<r_{g^\star}-r_g
\qquad\text{for every }g\neq g^\star,
\label{eq:chamber-safe-horizon}
\end{equation}
then $g^\star$ remains the unique active group at every prefix through step
$h$. Along such a path, the maximum reduces to one fixed group coordinate.
The statement is pathwise. It becomes a policy-level certificate when the
declared bounds cover every path reachable under that policy.
Appendix~\ref{app:uniform-details} derives the characterization and the horizon
claim.

For $\psi=\max$, Proposition~\ref{prop:uniform-control} gives $2\varepsilon$
regret on a finite menu. A selected candidate whose measured improvement over
a shared reference clears $2\varepsilon$ therefore has a lower true
worst-group endpoint. This resolves the current menu. Future increments and
descendant menus still depend on the chosen state, so a terminal guarantee
needs fresh control at each reached comparison or a trajectory-wide
continuation bound. Section~\ref{sec:moe-full-model} measures this distinction
at full-model scale.

Existing pruning methods occupy different points on this interface ladder.
Wanda reads pooled activation norms, SparseGPT adds a local Hessian, and
nonuniform rules add allocation signals
\citep{sun2024wanda,frantar2023sparsegpt,yin2024owl,xu2024besa,
choenni2025mwanda}. HOPE preserves Hilbert
geometry under one surrogate measure \citep{mobahi2026hope}. In MoE pruning,
static scorers aggregate pooled routing or activation signals and may
re-estimate them after deletion
\citep{lasby2025reap,liu2026scoreexperts,jaiswal2025fantastic};
domain-directed methods rank experts from a
target-domain route profile \citep{pei2026premo,jha2026halfexperts}; and
coalition-aware methods score routed co-occurrence \citep{zhang2026shape}.
Layerwise enumeration and full-model search make complete intervention queries
over restricted families, while global allocation methods redistribute expert
budgets across depth
\citep{lu2024notall,liu2024eep,zhang2026grape}. These methods improve how
experts are scored or searched. The interface question asks whether the
information consumed by that search licenses its groupwise endpoint decision.

Compatibility classes are related to identified sets in decision theory under
partial identification \citep{manski2008partial,yata2021optimal}. Here the
identified object is a candidate-by-group endpoint field. The
compression-specific contribution derives information prices and decision-grain
consequences for concrete pruning interfaces, then tests them across dense and
routed architectures.

The theory predicts a split. Group resolution can retain broad severity
contrasts that pooling may erase. Fine selection requires the pushforward
interval of its declared decision contrast to avoid zero. The experiments test
both predictions.

\section{Group resolution recovers broad severity}

Pooling can hide concentrated group damage. The empirical question is
whether exposing the group rows restores any endpoint-relevant order. Across a
32-condition Llama-3.2-3B-Instruct grid, the group-resolved diagonal ranked
realized worst-group damage with Spearman correlation 0.9239
\citep{llama32model}. The grid crosses Wanda and SparseGPT, four sparsity
settings, and four calibration treatments over equal-weight general, rare-code,
rare-knowledge, and safety-labeled sources
\citep{raffel2020t5,haskelldataset,mallen2023popqa,ji2025pkusaferlhf}.
Appendix~\ref{app:empirical-details} gives the data separation and matched
cross-model protocol.

\begin{figure}[t]
\centering
\includegraphics[width=\linewidth]{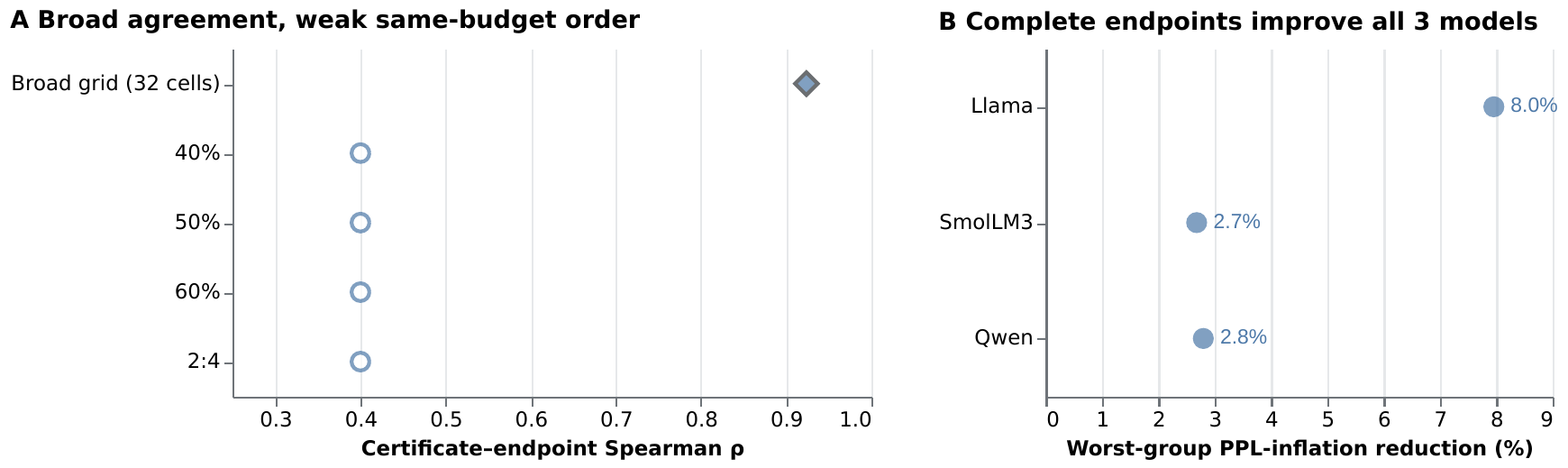}
\caption{Empirical overview of the dense-model boundary. Group-resolved local
evidence recovers broad severity, while target-matched complete endpoints
recover fine gains.
(A) The diagonal certificate ranks the full
32-cell grid with Spearman correlation 0.9239, while all four SparseGPT
same-budget correlations are 0.4 ($n=4$ treatments per budget). (B) Worst-group
PPL-inflation reductions relative
to each model's early-preserving 60\%-sparse reference; all paired excess-NLL
intervals lie below zero.}
\label{fig:main-results}
\end{figure}

For every grid cell we compare
\begin{equation}
\mathcal C_D(M)=\max_g\sum_{u:M_u=1}w_u^2n_g(u)^2
\end{equation}
with
\begin{equation}
\Delta_g^{\mathrm{PPL}}(M)=\frac{\operatorname{PPL}_g(M)}
{\operatorname{PPL}_g(\mathrm{dense})}-1,
\qquad
\Delta_{\max}^{\mathrm{PPL}}(M)=\max_g\Delta_g^{\mathrm{PPL}}(M).
\end{equation}
SparseGPT's certificate scores its mask before compensation.
Figure~\ref{fig:main-results} gives an empirical overview. Panel A shows the
severity comparison; Panel B previews the complete-endpoint results in
Section~\ref{sec:dense-selection}.

That broad success has a sharp boundary. All four SparseGPT within-budget
correlations were 0.4 ($n=4$ treatments per budget), and OBS compensation
improved 7 of 16 matched cells. At the broader calibration contrast,
group-balanced calibration reduced worst-group damage relative to pooled C4 by
21.7\% for Wanda and 31.5\% for SparseGPT. A matched-token Mistral-7B contrast
gave a 29.8\% reduction \citep{mistral7bv03model}.

The diagonal is therefore useful as a broad-severity diagnostic. Fine mask order
calls for evidence at the level of the complete intervention.

\section{Why local information still cannot decide endpoints}

\subsection{Exposure and ownership diverge}
\label{sec:exposure-ownership}

Exposure and restoration leverage are different observables. Group contrast
localized to down and output projections, while restoring those families lost
to an equal-count control at all four budgets.

\begin{figure}[t]
\centering
\includegraphics[width=0.95\linewidth]{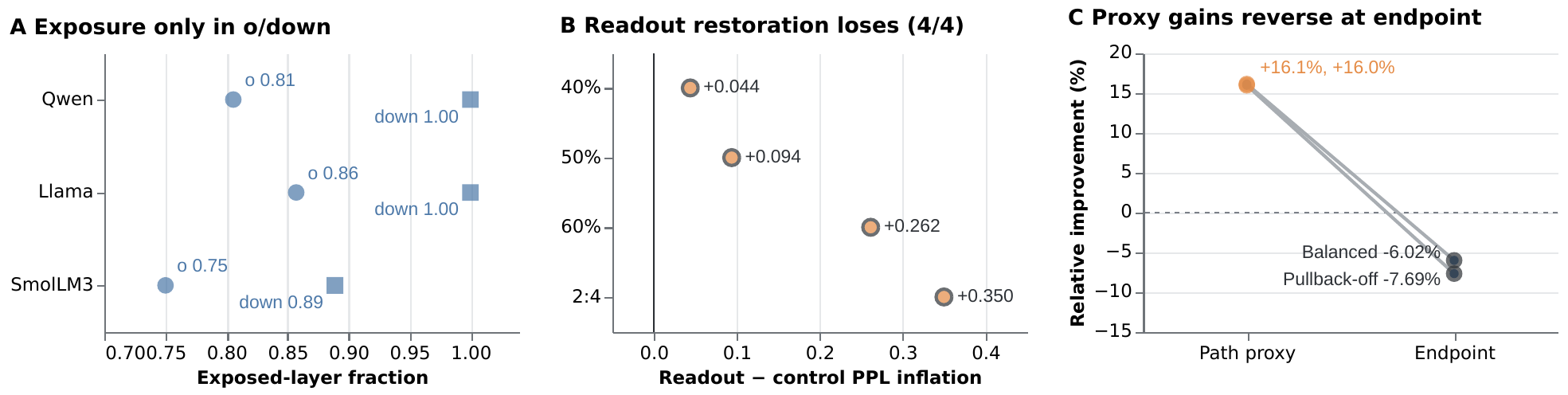}
\caption{Visibility does not determine leverage or endpoint order. (A) Exposure
occurs only in o/down projections across the three models. (B) Restoring those
families loses to the control at all four budgets. (C) Both path-proxy gains
reverse at the sparse endpoint.}
\label{fig:footprint}
\end{figure}

For activation channel $i$ with positive pooled second moment, define
\begin{equation}
\rho_i=\frac{\max_g n_g(i)^2}{G^{-1}\sum_g n_g(i)^2}.
\end{equation}
Under the fixed threshold in Appendix~\ref{app:empirical-details}, positives
occurred only in down and output projections on Llama and SmolLM3.
Figure~\ref{fig:footprint} also shows the Qwen profile.

The map locates exposure. Output and down projections consume attention-head
and FFN features, so their inputs show where group contrast becomes visible.
Ownership would require absolute mass and causal leverage, neither of which is
contained in the scale-invariant statistic $\rho$.

Both restoration arms improved perplexity. The control improved more each time
because it recovered 44--51 times more diagonal mass, even though down/output
restoration delivered 28--44 times more groupwise gain per unit mass. The
comparison identifies a mass--leverage decomposition. Because the arms use
different selection rules, it does not establish a total order over module
families. The definitions and selection asymmetry are in
Appendix~\ref{app:empirical-details}.

\subsection{A reliable path-quadratic score can reverse at the endpoint}
\label{sec:path-boundary}

Could a reproducible second-order score decide nearby sparse endpoints? An
empirical-Fisher quadratic proxy along the reference path reached 0.906
split-half reliability. Its predicted gains of 16.1\% and 16.0\% became
endpoint losses of 6.0\% and 7.7\% against the two controls. OBS-style
second-order information can therefore rank broad local severity while
miscalibrating distant endpoint order
\citep{singh2020woodfisher,wu2024iobs}.

Secondary diagnostics locate the gap more precisely. The full empirical-Fisher
form recovered 7 of 8 groupwise signs, yet a predicted 22.1\% gain shrank to an
inconclusive 2.8\%. By contrast, restricted actual-mask secants gave
actual-to-predicted ratios from 0.89 to 1.14 across 48 cells. The tested
family-wide norm-product bound was vacuous. Appendix~\ref{app:empirical-details}
gives the instrument and partition details.

\section{Complete dense endpoints recover target-matched gains}
\label{sec:dense-selection}

The boundary leaves a constructive route through finite menus of complete,
measured interventions. A deterministic generator
$\mathsf G:\Theta\to\mathcal M_s$ maps low-dimensional coordinates to a finite
menu $\mathcal Q$ of same-budget masks, where $\mathcal M_s$ is the set of
masks at budget $s$. Local evidence constructs the menu. Paired measurements
of the declared endpoint rank its members.

\subsection{Coarse structure transfers; fine decisions remain target-specific}

Coarse decision coordinates transferred across model families. The generator's
eight coordinates produce complete 60\% row-sparse masks, and its
early-preserving reference reduced worst-group
perplexity inflation by 20.9\%, 18.3\%, and 12.6\% on Llama, SmolLM3, and Qwen,
respectively, relative to balanced uniform allocation
\citep{llama32model,smollm3model,qwen25model}.
Appendix~\ref{app:empirical-details} defines the coordinates and menus.

The Qwen allocation contrast improved the safer-response margin by $0.0230$
(one-sided 95\% lower bound $0.0124$) and produced higher point estimates on
MMLU (37.3\% to 39.0\%) and PopQA answer match (10.3\% to 11.0\%)
\citep{ji2025pkusaferlhf,mallen2023popqa}. PopQA moved opposite to its
source-likelihood contrast, reinforcing that source likelihood alone does not
determine task order.

Fine directions were model and target specific. One forward-KL coordinate on
Llama reduced worst-group KL by 4.29\% and worst-group perplexity inflation by
7.96\%, with both paired intervals below zero. The corresponding Qwen and
SmolLM3 performance effects were inconclusive. Target-matched complete-menu
selection resolved the performance contrast. It lowered worst-group
perplexity inflation by 2.68\% on SmolLM3 and 2.80\% on Qwen, with paired
excess-NLL upper bounds below zero. Nonbinding groups could worsen while
remaining inside their available slack, as
Equation~\ref{eq:slack-adjusted-change} predicts.

Complete measurement removes transport ambiguity for each queried candidate.
Comparing many noisy measurements introduces a second cost, selection
pressure. A ten-mask stress test quantified that price. The winning mask
appeared 4.87\% better on the selection sample and 2.76\% worse on held-out
data; the absolute contrast moved from $-0.0782$ to $+0.0395$ (paired 95\%
interval $[0.0170,0.0635]$). This implies an empirical uniform-error floor of
$\varepsilon\geq0.0476$ for that selected comparison. Proposition
\ref{prop:uniform-control} therefore requires simultaneous control over the
menu or independent post-selection data.

Dense pruning makes complete-mask queries affordable by reducing the candidate
count. MoE lowers a different cost. Router traces attach group-resolved
information directly to the removable unit. In both cases, the remaining
question is whether local evidence survives composition into a complete
intervention.

\section{MoE exposes the same boundary at expert granularity}
\label{sec:moe-interface}

\subsection{The representation separates routing from expert function}

Dense layers hide the group decomposition behind shared activations. An MoE
router records part of that decomposition on the expert node that can be
removed. This makes MoE a constructive test of the same decision boundary.
The router lowers the cost of observing group structure, while the expert's
computed function remains unknown. Write a sparse MoE layer as
\begin{equation}
y(x)=\sum_{e=1}^{E}a_e(x)f_e(x),
\qquad a(x)=\mathsf{Route}(r(x)),
\label{eq:moe-representation}
\end{equation}
where $r(x)\in\R_+^E$ is the nonnegative gate trace and $\mathsf{Route}$ contains the
selection, normalization, and capacity rule. Router families differ through
$\mathsf{Route}$; the interface argument is unchanged
\citep{cai2025moesurvey,liu2026moeinference,qiu2025moebench}. At this routing
interface, a shared expert with equal group-mean coefficients occupies a
contrast-one column, the dense limit of the routing observation. Its computed
function still enters through the endpoint account below.

For each layer, group, and expert with positive mean route mass, define the
router-resolved exposure
\begin{equation}
A_{\ell e g}=\E_{x\sim P_g}[r_{\ell e}(x)],
\qquad
\rho_{\ell e}=
\frac{\max_g A_{\ell e g}}{G^{-1}\sum_g A_{\ell e g}}.
\label{eq:moe-exposure}
\end{equation}
For fixed nonnegative local costs $\kappa_{\ell e}$, setting
$u=(\ell,e)$, $A_{ug}:=A_{\ell e g}$, $e_{ug}=\kappa_uA_{ug}$, and
$\bar A_u=G^{-1}\sum_gA_{ug}$ in
Equation~\ref{eq:conic-pooling-identity} gives the exact relaxed pooled value
$G\sum_{u\in S}\kappa_u\bar A_u$. At fixed $|S|=k$, choosing the $k$ smallest
pooled route masses is therefore minimax-optimal for that exposure interface
when $\kappa_u=1$. It remains order-inconsistent for worst-group exposure. A
training constraint on pooled
route mass cannot determine the group rows of $A$.

Router exposure is attached to the expert node removed by the intervention,
avoiding the dense edge-to-node mismatch in
Section~\ref{sec:exposure-ownership}. The replacement supplied by surviving
experts remains an endpoint quantity.

\subsection{Router traces expose group structure at removable nodes}

\begin{figure}[H]
\centering
\includegraphics[width=\linewidth]{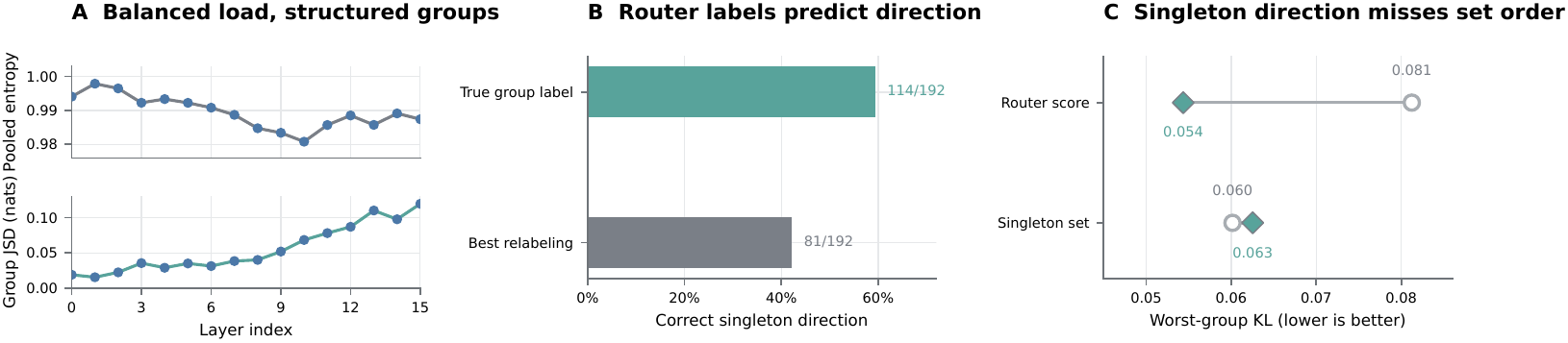}
\caption{Router traces expose group structure without resolving set order.
(A) Pooled load remains near uniform as group-row divergence grows. (B) Router
labels predict singleton direction in 114 of 192 deletions, versus 81 under the
strongest relabeling. (C) At 25\% deletion, group-resolved singleton selection
is worse by 0.00237 (one-sided upper bound 0.00686); the router-score pair is
descriptive.}
\label{fig:moe-router-audit}
\end{figure}

OLMoE provides a top-8-of-64 routing instance
\citep{muennighoff2025olmoe}. Its pooled load was nearly balanced, with entropy
from 0.981 to 0.998, while 253 of 1,024 expert-layer cells had group contrast
above 1.5. Group-row divergence also increased with depth. The structure was
reproducible. Router mass and routed contribution energy reached split-half
median Spearman correlations of 0.9947 and 0.9832, respectively
(Figure~\ref{fig:moe-router-audit}). At a fixed hidden state, stored router
logits determine post-deletion routing exactly. Later hidden-state changes
still require an endpoint query. Appendix~\ref{app:moe-proofs} gives the
derivation; Appendix~\ref{app:empirical-details} gives the measurement protocol.

Router logits exactly reproduced post-deletion routing for every measured set.
Deletion still changes the layer output and all later states. Singleton queries
measure one replacement, pair queries expose interaction, and complete-set
queries determine the intervention. Appendix~\ref{app:moe-proofs} gives the
conditional subadditive certificate.

\subsection{Endpoint queries resolve replacement and interaction}
\label{sec:moe-endpoint}

Router exposure predicted singleton direction. Across 192 singleton deletions,
it identified the maximizing group in 114 cases, compared with 81 for the
strongest relabeling control. That directional information did not compose
into set order. High co-routing had the larger median pair interaction in only
one of three layers, and at 25\% deletion neither singleton ranking nor
current-state singleton remeasurement separated the complete sets.

Complete-set queries resolved the two state-matched decisions. At the initial
state, the selected set reduced held-out worst-group forward KL by 13.7\%, with
a one-sided upper bound of $-0.00277$. At the resulting state, a measured
one-swap candidate produced a further 7.2\% sample reduction. Every terminal
group remained below the starting active-group mean; the largest adjusted
upper bound was $-0.000445$. The near tie between general and safety matches the
active-face view of the worst group as a state-dependent coordinate.

\begin{figure}[t]
\centering
\includegraphics[width=0.92\linewidth]{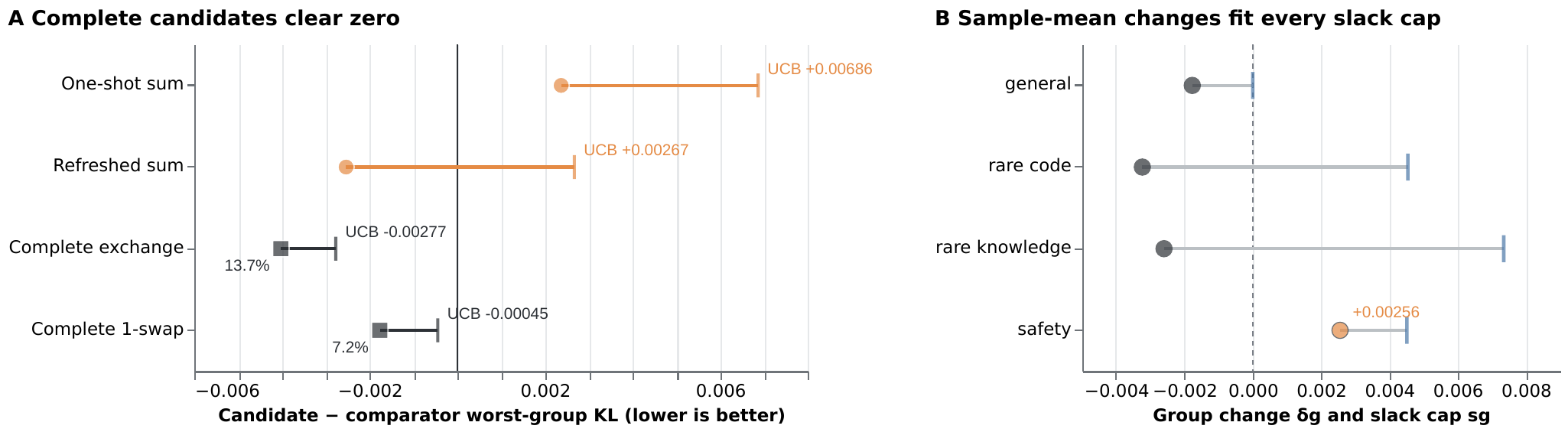}
\caption{Complete-candidate endpoints recover both set-level decisions missed
by singleton ranking. (A) Held-out worst-group KL falls by 13.7\% and 7.2\%;
caps are one-sided upper bounds. (B) At the sample means, each one-swap group
change lies within its available slack. One layer and one deletion budget are
shown.}
\label{fig:information-loop}
\end{figure}

Router traces narrow the finite expert menu, and complete-set queries identify
its order. A validated uniform transport bound or a sufficiently tight
subadditivity certificate could replace those queries over the family it
separates. Appendix~\ref{app:empirical-details} gives the normalized diagnostics
and uncertainty procedure.
Figure~\ref{fig:information-loop} contrasts the failed singleton-derived
decision with the two move-matched endpoint decisions.

\subsection{Full-model composition separates refresh from hard max}
\label{sec:moe-full-model}

To test composition across depth, four rounds removed 16 of 64 experts from
each of OLMoE's 16 routed layers. Five current-state scores proposed candidates;
compute-matched trajectories evaluated the resulting proposals by pooled or
worst-group forward KL. Their first menu was shared. Later menus became
state-dependent after the trajectories diverged.

\begin{figure}[t]
\centering
\includegraphics[width=0.90\linewidth]{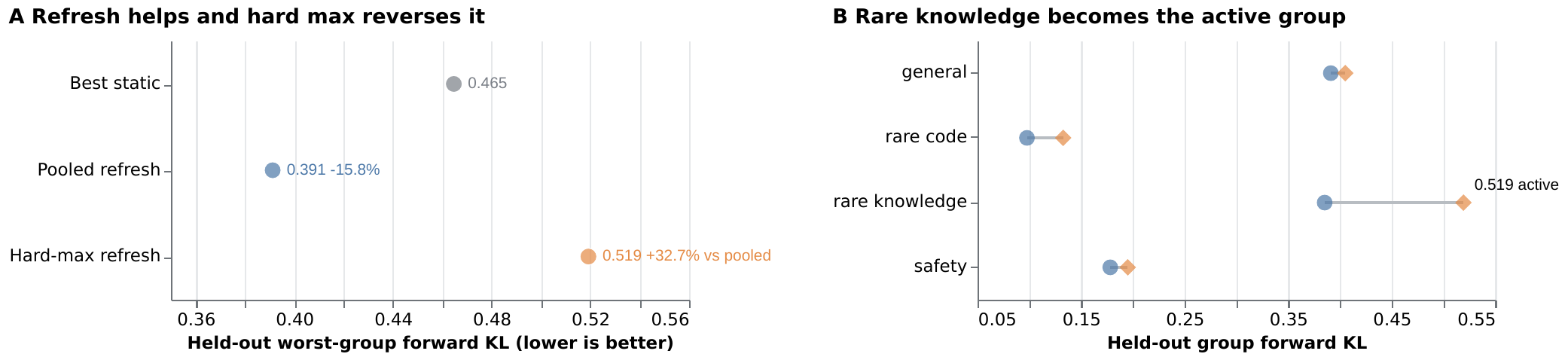}
\caption{Full-model composition. (A) Pooled refresh lowers held-out worst-group
forward KL by 15.8\% relative to the best static score; compute-matched hard
max is 32.7\% worse. (B) Hard max raises all four group risks and makes rare
knowledge active.}
\label{fig:full-model-moe}
\end{figure}

As Figure~\ref{fig:full-model-moe} shows, pooled refresh lowered held-out
worst-group KL from 0.4647 for the best static scorer to 0.3912, a 15.8\%
reduction with one-sided upper difference $-0.0356$. Hard max reached 0.5193,
32.7\% above pooled. Its paired difference was $+0.1281$, with a one-sided
95\% lower bound of $+0.0823$. Worst-group excess NLL
was 0.3157 for static REAP, 0.3655 for pooled refresh, and 0.3764 for hard max;
neither adaptive trajectory carried the KL gain to this performance endpoint.

At the final search endpoint, hard max lay only 0.0012 below pooled on the
measured maximum while raising pooled KL by 0.0346 and rare-knowledge KL by
0.0798. Its top-two gap contracted from 0.0876 to 0.0067, placing it about 13
times closer to an active-group chamber wall. From search to held-out data,
rare knowledge rose relative to general by 0.0817 under pooled selection and
stayed inside the 0.0876 margin. Under hard max it rose by 0.1208, crossed the
0.0067 margin, and became active. All 64 rare-knowledge sequence differences
favored pooled. The pooled margin absorbed the relative shift, while the
hard-max trajectory crossed an active-group chamber wall. Across search rounds,
each selected mask also changed the candidates available at the next state.
Endpoint refresh therefore improved a static score in this trajectory, while
stepwise hard-max minimization did not compose. A terminal policy must control
both the evolving active face and the menus generated along its path.

\section{Discussion}

Retain group decomposition until complete candidates are affordable, then carry
the slack profile after each accepted move. Dense generators reduce the number
of candidates. MoE routers attach group-resolved evidence to removable experts.
A chamber-safe margin supports several steps only while declared drift bounds
cover reachable paths. When the margin shrinks, refresh the endpoint state or
compare nearby active faces. In OLMoE, hard max finished near a chamber wall and
the held-out shift crossed it. Per-group caps therefore protect one measured
step. Longer horizons require fresh comparisons or a continuation bound. Coarse
depth allocation transferred across three dense models, while fine choices
changed with response secants, active group, and slack.

\paragraph{Limitations.}
The exact pooled and capped factors hold for any positive linear functional on
a convex cone, including full PSD input second moments for a fixed candidate.
The reusable per-weight score, switching relaxation, and
experiments use diagonal layer reconstruction in a fixed basis. Cross-layer
transport requires blockwise Lipschitz bounds on the compared trajectories and
a terminal Lipschitz or Hessian bound. The tested spectral product was vacuous
at both depths. A useful exceptional-set bound would have to hold under the
perturbation-energy measure and remain valid after persistent high-gain
directions are removed. We have not established such a tail bound for trained
networks.

The dense evidence covers three model families at one main sparsity budget and
the declared candidate menus. Task bridges evaluate one Qwen mask contrast and
do not replace a broad behavioral benchmark. The MoE evidence covers one
checkpoint, one-layer finite menus, and one 16-layer trajectory at 25\%
deletion. Reusing the search sample across rounds leaves trajectory structure
and adaptive estimation error entangled. No task-level MoE gain is established.

\section{Conclusion}

Group-resolved local evidence recovered broad severity and generated tractable
candidates. Complete endpoint measurements were still needed to rank the
queried dense and MoE interventions. The full-model trajectory added a second
state variable, the slack to the active-group chamber walls. Pooled refresh improved
teacher KL, while greedy hard max crossed a wall; neither trajectory improved
excess NLL. Local statistics propose, fresh endpoint measurements decide, and
sequential policies refresh state before their slack margin expires.

\section*{Broader impact statement}

Group-resolved measurements may detect regressions hidden by aggregate calibration.
The source groups in this study are stress-test partitions, and their NLLs do
not establish capability, safety, or demographic fairness. Deployment claims
require task-specific behavioral evaluation. Compression can degrade rare or
safety-critical behavior even when average loss improves. A deployment should
therefore use group-relevant safety tests, retain an uncompressed rollback, and
set rejection thresholds before release.

\section*{Reproducibility statement}

All definitions, assumptions, and proofs needed for the theoretical claims
appear in the paper. Appendix~\ref{app:empirical-details} records the data
separation, fixed candidate menus, and evaluation procedures used in the
empirical study.

\bibliography{references}
\bibliographystyle{tmlr}

\appendix

\section{Proofs for the interface results}
\label{app:interface-proofs}

\subsection{The switching relaxation}
\label{app:switching}

For nonnegative row damage $e_{ug}$, define
\begin{equation}
\Phi_D(S)=\max_g\sum_{u\in S}e_{ug},
\qquad
\Psi_D(S)=\sum_{u\in S}\max_g e_{ug}.
\label{eq:switching}
\end{equation}
Then $\Phi_D(S)\leq\Psi_D(S)$. At fixed cardinality, sorting the smallest
$\max_g e_{ug}$ values minimizes $\Psi_D$. Equality holds exactly when one
group maximizes every selected row, allowing ties. The difference
$\Psi_D(S)-\Phi_D(S)$ is the switching premium.

\paragraph{Proof of Proposition~\ref{prop:matched-reversals}.}
For the pooled witness, take two units with profiles $(4,0,0,0)$ and
$(1,1,1,1)$. Exchanging the profiles preserves each pooled mean and reverses
their worst-group values. For the downstream witness, take one group,
$W=\Sigma=I_2$, and let the two masks remove opposite output coordinates.
Their error matrices are $\diag(1,0)$ and $\diag(0,1)$; the pullbacks
$\diag(\Lambda,\mu)$ and $\diag(\mu,\Lambda)$ reverse the order while leaving
the input-side observation fixed. PSD self-duality gives the stated robust
dominance condition. Finally, compactness lets us choose disjoint open balls
around $a$ and $b$ whose closures miss $\Gamma$. Nonnegative smooth bumps
supported in those balls vanish with every derivative on $\Gamma$. Exchanging
the bumps reverses the endpoint order.

\paragraph{Cross-layer propagation under declared Lipschitz bounds.}
Let $h_{k+1}=T_k(h_k)$ and
$\widetilde h_{k+1}=\widetilde T_k(\widetilde h_k)$ start from the same input,
and suppose $T_k$ is $a_k$-Lipschitz on a set containing $h_k$ and
$\widetilde h_k$ for every evaluated input. Define the
on-trajectory block discrepancy
$r_k=T_k(\widetilde h_k)-\widetilde T_k(\widetilde h_k)$ and
$D_{kg}=\E_g\|r_k\|_2^2$. The recursion
\begin{equation}
\|h_{k+1}-\widetilde h_{k+1}\|_2
\leq a_k\|h_k-\widetilde h_k\|_2+\|r_k\|_2
\end{equation}
and Minkowski's inequality give
\begin{equation}
\left(\E_g\|h_L-\widetilde h_L\|_2^2\right)^{1/2}
\leq
\sum_{\ell=0}^{L-1}
\left(\prod_{k=\ell+1}^{L-1}a_k\right)\sqrt{D_{\ell g}}.
\label{eq:multilayer-propagation}
\end{equation}
For $T_k=I+F_k$ with $F_k$ $\beta_k$-Lipschitz, one may take
$a_k=1+\beta_k$; the single-error bound is obtained by retaining only its
$\ell$th summand and squaring.
Let $B_g$ denote the right-hand side of
Equation~\ref{eq:multilayer-propagation}. If the per-sample terminal loss
$\phi_g(\cdot;x)$ is $\ell_g^{\mathrm{out}}$-Lipschitz on the segment joining $h_L(x)$ and
$\widetilde h_L(x)$, uniformly over evaluated inputs, then
\begin{equation}
\left|\E_g[\phi_g(h_L;x)-\phi_g(\widetilde h_L;x)]\right|
\leq \ell_g^{\mathrm{out}} B_g.
\end{equation}
If instead $\nabla_z\phi_g(h_L(x);x)=0$ for every evaluated input and the
Hessian operator norm is at most $L_g^{(2)}$ on each joining segment, Taylor's
theorem gives the quadratic bound $L_g^{(2)}B_g^2/2$. These are sufficient
continuation assumptions; the dense-point diagonal moments used in the study
do not supply them.

For a fixed group, candidate perturbation $\delta(x)$, and suffix Jacobian
$J(x)$, set $e(x)=\|\delta(x)\|_2^2$. Let $P(x)$ be a measurable orthogonal
spectral projector of $J(x)^*J(x)$ onto a high-gain subspace. Define
$Z(x)=\|P(x)\delta(x)\|_2^2/e(x)$ when $e(x)>0$ and zero otherwise, and let
$r(x)=\|J(x)\delta(x)\|_2^2$. If $\sigma_1$ and $\sigma_{\mathrm{tail}}$
uniformly bound the singular values on $\operatorname{range}P(x)$ and its
orthogonal complement, respectively, spectral orthogonality gives the mixture
bound used below. The next lemma converts a tail bound for $Z$ into an average
transport bound.

\begin{lemma}[Energy-weighted exceptional-set closure]
\label{lem:energy-weighted-closure}
Let $e\geq0$ satisfy $0<\bar e:=\E_g e<\infty$, and define the probability
measure $d\nu_g=e\,dP_g/\bar e$. Let $0\leq Z\leq1$, choose
$0<\beta<1$, and set $a_\beta=(1-\beta)^2$. If
\begin{equation}
\nu_g\{Z\geq a_\beta\}\leq p_\beta\leq1,
\end{equation}
then
\begin{equation}
\E_g[eZ]\leq \tau_{\mathrm{eff}}\bar e,
\qquad
\tau_{\mathrm{eff}}
=a_\beta+(1-a_\beta)p_\beta.
\label{eq:energy-weighted-cap}
\end{equation}
In particular, $p_\beta<1$ implies $\tau_{\mathrm{eff}}<1$. If
$0\leq\sigma_{\mathrm{tail}}\leq\sigma_1$ and
\begin{equation}
r(x)\leq e(x)\{Z(x)\sigma_1^2+(1-Z(x))\sigma_{\mathrm{tail}}^2\},
\end{equation}
then
\begin{equation}
\E_g r\leq
\{\tau_{\mathrm{eff}}\sigma_1^2
+(1-\tau_{\mathrm{eff}})\sigma_{\mathrm{tail}}^2\}\bar e.
\label{eq:energy-weighted-spectral}
\end{equation}
\end{lemma}

Indeed, if $q=\nu_g\{Z\geq a_\beta\}$, splitting the expectation gives
$\E_{\nu_g}Z\leq q+a_\beta(1-q)\leq\tau_{\mathrm{eff}}$; monotonicity of the
spectral mixture yields Equation~\ref{eq:energy-weighted-spectral}. Energy
weighting is essential because a set with small $P_g$ probability can carry
all perturbation energy. If the perturbation lies in the maximal-gain subspace
$\nu_g$-almost surely, then $p_\beta=1$ and the bound gives no strict
improvement.

\subsection{Transport and shared-reference consequences}
\label{app:uniform-details}

For contexts $c,c'$ with matched group indices and endpoint scales,
Equation~\ref{eq:slack-adjusted-change} and the 1-Lipschitz property of the
maximum give
\begin{equation}
|\Delta_F^c-\Delta_F^{c'}|
\leq
\|(\delta^c-s^c)-(\delta^{c'}-s^{c'})\|_\infty
\leq \|\delta^c-\delta^{c'}\|_\infty
+\|s^c-s^{c'}\|_\infty.
\label{eq:active-face-transport}
\end{equation}

Proposition~\ref{prop:uniform-control} already gives the finite-menu selection
bound. If the same simultaneous error also covers a shared reference
$M_{\mathrm{ref}}$, the triangle inequality gives
\begin{equation}
|\Delta_F(\widehat M;M_{\mathrm{ref}})
-\widehat\Delta_F(\widehat M;M_{\mathrm{ref}})|\leq2\varepsilon.
\label{eq:finite-menu-certificate}
\end{equation}
Thus a measured shared-reference improvement that clears $2\varepsilon$
certifies the true order.

\paragraph{Derivation of the response-state characterization.}
Equation~\ref{eq:slack-adjusted-change} follows by writing
$r_g=m(r)-s_g(r)$. If $s(r)=s(r')$, the equation gives identical response
functions. Conversely, $q_r(s(r))=0$. Equality of the two response functions
therefore gives
$\max_g\{s_g(r)-s_g(r')\}=0$, so $s(r)\leq s(r')$ coordinatewise. Exchanging
$r$ and $r'$ gives the reverse inequality. Finally, equal slack implies
$r'_g-r_g=m(r')-m(r)$ for every group, while adding a common constant changes
the maximum by the same constant and preserves slack. Equation
\ref{eq:slack-state-update} follows from
$m(r+\xi)=m(r)+q_r(\xi)$.

For the horizon claim, summing
$\xi_{t,g}-\xi_{t,g^\star}\leq b_{t,g}$ over any prefix and applying
Equation~\ref{eq:chamber-safe-horizon} keeps $g^\star$ strictly above every
other group.

\paragraph{Proof of Proposition~\ref{prop:uniform-control}.}
The Lipschitz condition and Equation~\ref{eq:uniform-interface-error} give
$|f(M)-\widehat f(M)|\leq L_\psi\varepsilon$ for every queried mask. This
gives both threshold inclusions directly. For selection,
$f(\widehat M)\leq\widehat f(\widehat M)+L_\psi\varepsilon
\leq\widehat f(M^*)+L_\psi\varepsilon
\leq f(M^*)+2L_\psi\varepsilon$ for an exact minimizer $M^*$.

\paragraph{Local sufficient radius.}
For the KL specialization $F_g=J_g$, regard $J_g$ as a function of student
parameters and write
$\theta_M=\theta_0+\Delta_M$ with $\|\Delta_M\|_2\leq r$. Suppose an interface
supplies $\widehat b_g$ and self-adjoint $\widehat H_g$ such that
\begin{equation}
\|\nabla J_g(\theta_0)-\widehat b_g\|_2\leq a_g,
\qquad
\|\nabla^2J_g(\theta_0)-\widehat H_g\|_{\mathrm{op}}\leq d_g,
\end{equation}
and the Hessian is $\Lambda_g^{\mathrm{Lip}}$-Lipschitz on every candidate
segment. Define
\begin{equation}
\widehat J_g(M)=J_g(\theta_0)
+\langle\widehat b_g,\Delta_M\rangle
+\frac12\langle\Delta_M,\widehat H_g\Delta_M\rangle.
\end{equation}
Then Equation~\ref{eq:uniform-interface-error} holds with
\begin{equation}
\varepsilon(r)=\max_g\left(
a_gr+\frac{d_g}{2}r^2+\frac{\Lambda_g^{\mathrm{Lip}}}{6}r^3\right).
\label{eq:local-filtration-radius}
\end{equation}
At dense teacher equality the first-order KL term vanishes and the Hessian is
the Fisher pullback; at a generic sparse iterate the first-order term remains.

Second-order Taylor expansion, Cauchy--Schwarz, and the stated operator and
Hessian-Lipschitz bounds give Equation~\ref{eq:local-filtration-radius}.

\section{Proofs for the MoE interface}
\label{app:moe-proofs}

\paragraph{Derivation of fixed-state rerouting.}
For router logits $z_{\ell e}(h)$, let
$\pi_{\ell e}(h)=\exp z_{\ell e}(h)/\sum_j\exp z_{\ell j}(h)$ and
$T_\ell(h)=\operatorname{TopK}_8\{z_{\ell e}(h)\}$. In this instance,
$r_{\ell e}(h)=\pi_{\ell e}(h)$ in Equation~\ref{eq:moe-representation}, and
$M_S$ denotes deletion of $S$ with at least eight surviving experts. If
$m_{\ell S}(h)=\sum_{s\in S}\pi_{\ell s}(h)$, deletion before the full softmax
gives
\begin{equation}
\pi^S_{\ell e}(h)=\frac{\pi_{\ell e}(h)}{1-m_{\ell S}(h)},
\qquad
T^S_\ell(h)=\operatorname{TopK}_8\{z_{\ell e}(h):e\notin S\}.
\label{eq:fixed-state-rerouting}
\end{equation}
Removing the deleted exponential terms gives
Equation~\ref{eq:fixed-state-rerouting}; the ratio
$\pi^S_{\ell e}/\pi^S_{\ell f}=\exp(z_{\ell e}-z_{\ell f})$ preserves the
strict order and equality classes under the fixed tie rule.

\paragraph{Derivation of Equation~\ref{eq:moe-set-certificate}.}
For deletion set $S$, abbreviate $J_g(S)=J_g(M_S)$ and let
$J_{ug}=J_g(\{u\})$. If $J_g(S)\leq\sum_{u\in S}J_{ug}$ for every group, then
\begin{equation}
\max_g J_g(S)
\leq \max_g\sum_{u\in S}J_{ug}
\leq \sum_{u\in S}\max_gJ_{ug}.
\label{eq:moe-set-certificate}
\end{equation}
This is groupwise subadditivity followed by the switching inequality in
Equation~\ref{eq:switching}.

\section{Empirical protocols and diagnostics}
\label{app:empirical-details}

\subsection{Common data and inference protocol}

Calibration, candidate selection, and final evaluation use disjoint source
rows unless a comparison is explicitly descriptive. For worst-group endpoints,
bootstrap resampling is performed within each group and the maximizing group is
recomputed in every draw. Task comparisons use paired row resampling.

The source partition was checked on packed-block halves and against 10,000
label permutations and 10,000 size-matched random-C4 partitions. Because the
halves reuse source rows, this comparison measures reliability.
The 7B comparison first matches average perplexity by linear interpolation and
then compares worst-group damage.

\subsection{Dense-model measurements}

An exposure instance is positive when at least 5\% of channels satisfy
$\rho_i\geq2$; a family is positive when at least half its instances are
positive. Figure~\ref{fig:footprint} reports the complete family counts. The
restoration comparison is selection-asymmetric. One arm restores an entire
module family, while its equal-count control draws from the global top-energy
tail. With
$\Delta D_{g,a}=D_g(\mathrm{base})-D_g(a)$ and
$\Delta P_{g,a}=\Delta_g^{\mathrm{PPL}}(\mathrm{base})-
\Delta_g^{\mathrm{PPL}}(a)$, we report the intervention secant
$\lambda_{g,a}=\Delta P_{g,a}/\Delta D_{g,a}$ when
$\Delta D_{g,a}>0$.

For the path-quadratic diagnostic, let $x_{\ell ti}(\tau)$ denote the module
input and $g_{\ell to}(\tau)$ the observed-token NLL gradient at
$\tau\in\{0,1/2\}$. We use
\begin{equation}
m_{\ell oi,g}(\tau)=\E_g\!\left[\operatorname{mean}_t
g_{\ell to}(\tau)^2x_{\ell ti}(\tau)^2\right],\qquad
q^{\mathrm{path}}_{\ell oi,g}=w_{\ell oi}^2
\left\{\tfrac13m_{\ell oi,g}(0)+\tfrac23m_{\ell oi,g}(1/2)\right\}.
\end{equation}
This is an empirical-Fisher proxy. It equals the corresponding KL Fisher only
after expectation over teacher-distributed targets at teacher-student equality.
The quadrature weights integrate $2(1-\tau)$ exactly for quadratic evolution.
Transport and split-half reliability are output-row Spearman correlations over
72 late-module-by-group cells. Mask fitting, proxy evaluation, and endpoint
perplexity use disjoint partitions.

The suffix-transport check starts from two complete 60\%-sparse SmolLM3 masks.
At six depths, one block's pruned projections are restored to one-half and full
dense magnitude while all other blocks remain masked. Hidden-state and terminal
logit RMS secants are recorded by group. Gains fitted on the first mask predict
the second without another fitted constant. The same masked suffix supplies the
spectral norm-product comparison in Equation~\ref{eq:multilayer-propagation}.

The dense generator has five centered sparsity coordinates over three depth
bands and two module families, plus three logits for a four-source calibration
mixture. Each coordinate produces a complete 60\% rowwise mask. Two fixed
24-mask Sobol menus contain 47 unique masks. Development uses the maximum group
change over three disjoint blocks and requires improvement on every block. The
selected candidate is evaluated on source-disjoint rows. Model-family
comparisons reuse the same menu and selection rule around the corresponding
reference mask.

The active-face screen additionally requires
$\delta_g\leq\tfrac12s_g$ for every group on the full development sample and
each block. Confirmation evaluates the selected generator coordinate on 96
new sequences per group. Each model realizes that coordinate as its own packed
mask.
Compression excess $E_g$ and forward KL $J_g$ are measured separately. The
Qwen performance contrast holds structure fixed and evaluates four prespecified
interpolations of the calibration mixture on 20 new sequences per group.

Behavioral checks use fixed mask pairs. PKU-SafeRLHF reports the paired change
in safer-response log-probability margin on 512 rows
\citep{ji2025pkusaferlhf}. MMLU uses all 14,042 test rows with five development
examples per subject, and PopQA uses all 2,852 held-out rows with the author's
alias-match score \citep{mallen2023popqa}. Their intervals come from paired row
resampling. The SmolLM3 confirmation uses 96 new sequences per group and draws
its rare-knowledge component from EntityQuestions
\citep{sciavolino2021entityquestions}.

\subsection{MoE measurements}

The router analysis uses 64 length-1024 sequences per group across all 16 routed
layers. Let $\bar A_{\ell e}=G^{-1}\sum_gA_{\ell eg}$ and define
\begin{equation}
p^{\mathrm{pool}}_{\ell e}=\frac{\bar A_{\ell e}}{\sum_j\bar A_{\ell j}},
\qquad
H_\ell=-\frac{\sum_e p^{\mathrm{pool}}_{\ell e}
\log p^{\mathrm{pool}}_{\ell e}}{\log E}.
\end{equation}
For group rows, $p_{\ell ge}=A_{\ell eg}/\sum_jA_{\ell jg}$ and
$m_{\ell e}=G^{-1}\sum_gp_{\ell ge}$. We report
$\mathrm{JSD}_\ell=G^{-1}\sum_g\KL(p_{\ell g}\|m_\ell)$. Routed contribution
energy for selected set $T_\ell$ is
\begin{equation}
Q_{\ell e g}=\E_{x\sim P_g}\left\|
\mathbf1\{e\in T_\ell\}\pi_{\ell e}f_{\ell e}(h)
\right\|_2^2.
\label{eq:moe-q}
\end{equation}

Singleton and pair measurements use a separate 16-sequence-per-group pack in
layers 2, 8, and 14. Their normalized diagnostics are
\begin{equation}
\widetilde J_{\ell eg}=
\frac{J_{\ell eg}}{64^{-1}\sum_{e'=1}^{64}J_{\ell e'g}},\qquad
\widetilde I_{\ell efg}=
\frac{|J_{\ell,\{e,f\},g}-J_{\ell eg}-J_{\ell fg}|}
{J_{\ell eg}+J_{\ell fg}}.
\label{eq:moe-normalized-diagnostics}
\end{equation}
The router label is $\arg\max_gA_{\ell eg}$, while the singleton label is
$\arg\max_g\widetilde J_{\ell eg}$. The control takes the best nonidentity
permutation of the four router labels. Pair analysis contrasts the largest and
smallest pooled co-routing values
$C_{\ell ef}=\E[\mathbf1\{e,f\in T_\ell(x)\}]$ in each measured layer.

Static set-level arms delete 16 of 64 experts in layers 2, 8, and 14 and are
evaluated on a separate pack. They rank pooled utilization,
$\max_gQ_{\ell eg}$, pooled singleton KL, or worst-group singleton KL. The
one-layer refresh comparison remeasures singleton endpoints after each
four-expert deletion and uses fresh round packs plus a 64-sequence-per-group
final pack.

The three-target library uses separate search and held-out packs. Search
controls the three candidate contrasts with adjusted one-sided bounds. Nested
one-, two-, and four-swap masks use distinct proposal, selection, and evaluation
samples. Eligibility requires a worst-group KL reduction of at least 5\% with a
one-sided upper bound below zero. Final simultaneous contrasts compare every
terminal group with the start-general anchor. Negativity of all four upper
bounds places the terminal group maximum below the starting maximum.

The full-model comparison covers all 16 routed layers. Four rounds delete four
experts per layer in each round. Frequency, MAN, MSAN, actual-weight REAP, and
normalized REAP are recomputed at the current state and propose complete next
masks. Pooled and worst-group selectors receive the same five proposer families
and the same endpoint-query budget. Search uses 32 length-512 sequences per
group, and final evaluation uses an independent 64 per group. The final
bootstrap recomputes both the active group and the strongest static baseline in
every draw.

\end{document}